\documentclass[journal]{IEEEtran}
\usepackage{mathrsfs} 
\usepackage{amsmath,amsfonts,amssymb}
\usepackage{tabularx}
\usepackage{booktabs}
\usepackage{multirow}
\usepackage{cleveref}
\Crefname{table}{Table}{Table}
\usepackage{mathrsfs}
\usepackage{algorithmic}
\usepackage{algorithm}
\usepackage{array}
\usepackage[caption=false,font=normalsize,labelfont=sf,textfont=sf]{subfig}
\usepackage{textcomp}
\usepackage{stfloats}
\usepackage{url}
\usepackage{verbatim}
\usepackage{graphicx}
\usepackage{cite}
\usepackage{xcolor} 
\begin{document}
\title{Joint Distribution Alignment for Universal Domain Adaptation}

\author{Shizhe Li, Hongshan Pu, Mengying Xie, Yi Xiang, and Xiaowei Yang
\thanks{Shizhe Li, Yi Xiang, and Xiaowei Yang are with the School of Software
Engineering, South China University of Technology, Guangzhou 510006,
China (e-mail: shizheli98@163.com).}
\thanks{Hongshan Pu is with the School of Future Technology, South China University of Technology, Guangzhou 511442,
China}
\thanks{Mengying Xie is with the College of Computer Science, Chongqing University, Chongqing 400044,
China}}

\markboth{}%
{Li \MakeLowercase{\textit{et al.}}: JOINT DISTRIBUTION ALIGNMENT FOR UNIVERSAL DOMAIN ADAPTATION}


\maketitle

\begin{abstract}
Unsupervised domain adaptation (UDA) has been widely concerned in the fields of machine learning, pattern recognition, and computer vision. Traditional UDA learning usually assumes that the label spaces of the source and target domains are exactly the same and only needs to solve the problem of sample distribution drift existing between two domains. However, in real world applications, the label spaces between two domains may be different. In this case, there are both sample distribution drift and class spatial difference between domains, namely Universal Domain Adaptation (UniDA) learning scenario. At present, existing works rarely offer theoretical analysis for universal domain adaptation. In this paper, we provide an upper bound of the generalization error for universal domain adaptation. According to the proposed generalization error bound, we propose a novel UniDA algorithm called Joint Distribution Alignment for Universal Domain Adaptation (JAUA), which aligns the joint distributions by minimizing the distribution discrepancy calculated by Chi-Square divergence. Furthermore, we propose a progressive pseudo-labeling method to assign the pseudo labels to unlabeled target samples. The experiment results on six public image datasets demonstrate the superiority of JAUA in handling the UniDA problem.
\end{abstract}


\section{Introduction}
\begin{figure}[!t]
\centering
\includegraphics[width=3in]{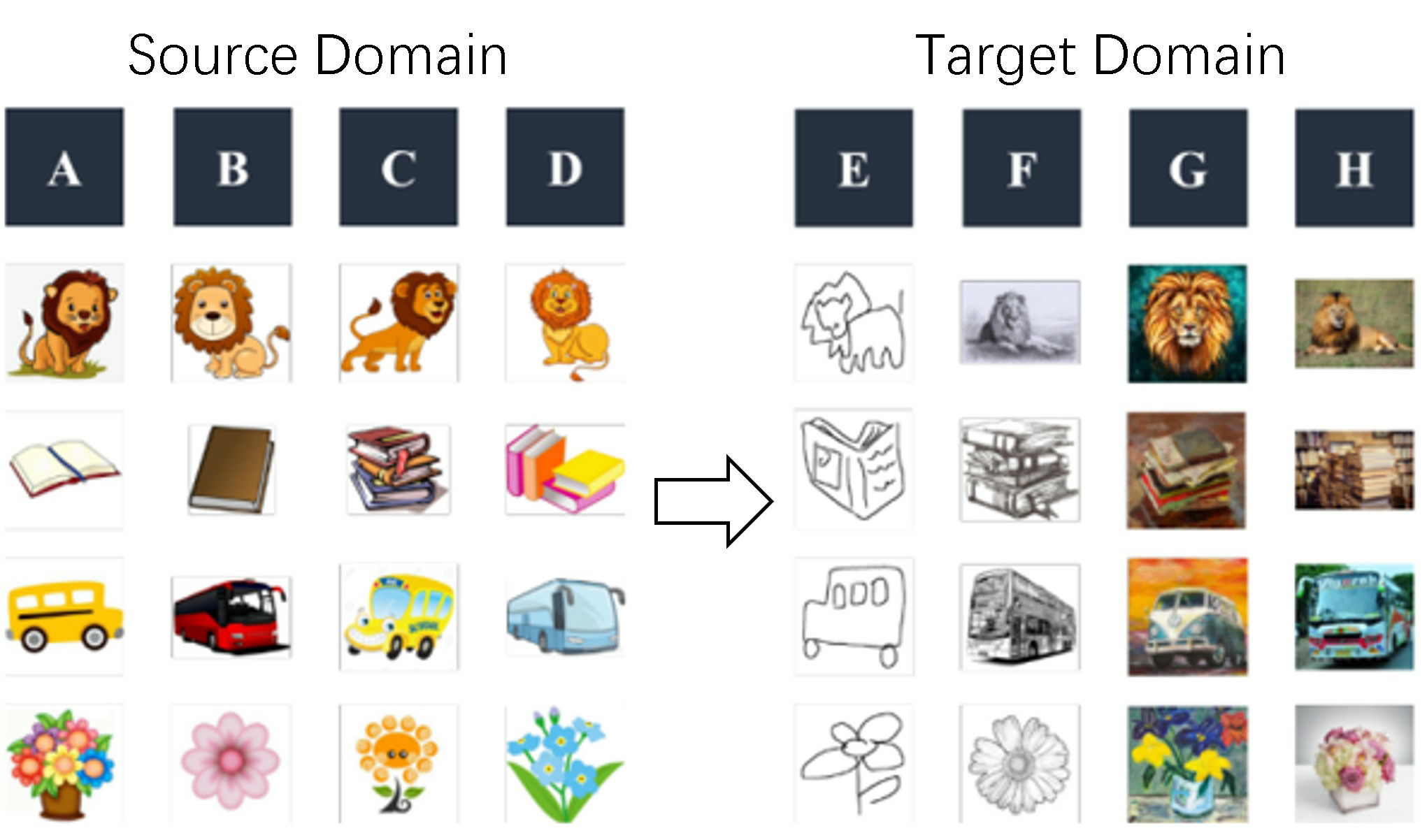}
\caption{Illustration of DA}
\label{fig1}
\end{figure}

\IEEEPARstart{T}{raditional} machine learning follows the basic assumption that both the training data and the test data follow the same joint probability distribution \cite{Pan2010}. However, in practical applications such as biological imaging, medical imaging, and industrial defect detection, the data distribution may shift due to environmental conditions, camera angles, or device status, making it challenging to satisfy this identical distribution assumption \cite{zhang2018task}. Under such conditions, classifiers with excellent performance on source images cannot be directly adapted to target image classification tasks. For efficiency reasons, transferring a trained model in one domain to another can save training costs significantly. To maintain the model performance in new domains, it is often necessary to recollect and relabel training data, which incurs significant human, material, and time costs. To address this issue, transfer learning treats the training data as the source domain and the test data as the target domain, aiming to identify transferable or reusable features, parameters, or models across domains. As a critical category in transfer learning, domain adaptation (DA) focuses on scenarios where the data distributions in the source and target domains are inconsistent, known as domain shift \cite{zhang2013domain,Niall2010Dataset}, as shown in Fig. \ref{fig1}.

\begin{figure}[!t]
\centering
\includegraphics[width=3in]{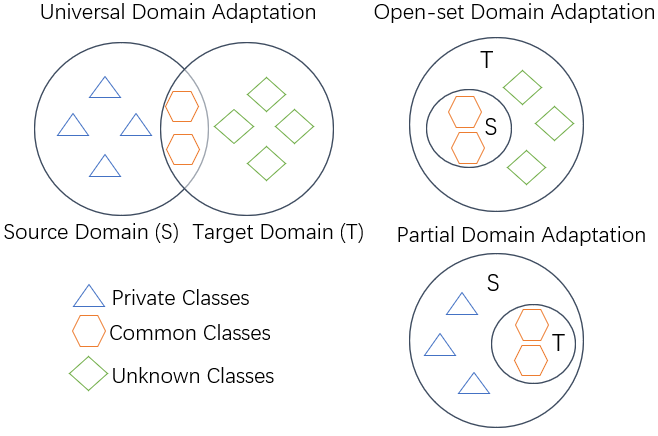}
\caption{Different setting of DA}
\label{fig2}
\end{figure}

Unsupervised domain adaptation (UDA) improves the model performance on unlabeled target domain by using knowledge learned in labeled source domain \cite{ben2010theory,dutta2026pose,liu2024universal}. Traditional unsupervised domain adaptation methods assume that both domains share the same label set. However, in real-world scenarios, since the data of the target domain are usually unlabeled, the difference of label spaces between domains cannot be predicted \cite{gao2020adversarial,ren2026open}. To tackle this limitation, the Universal Domain Adaptation (UniDA) framework has been proposed. In UniDA, both the source and target domains may contain exclusive classes, as shown in Fig. \ref{fig2}. In this way, UniDA can be closer to practical application and has higher practicality and scalability. At present, UniDA has been widely applied in many fields such as object detection \cite{zheng2025universal}, semantic clustering \cite{he2025target}, time series analysis \cite{mussard2025deep}, remote sensing image classification \cite{guo2024c3da,xu2023universal} and machine fault diagnosis \cite{qian2024adaptive}.

Under the UniDA framework, two key challenges must be addressed: 1) Label space discrepancy: identifying and labeling target-domain-specific classes as "unknown" while correctly classifying the common classes. 2) Distribution shift: aligning the joint distributions of the common classes across domains. 

To address the first challenge, this study uses a progressive separation mechanism based on the idea of \cite{liu2019separate}, which gradually distinguishes the common class samples from unknown class samples by selecting samples with higher confidence as common class samples. To address the second challenge, the upper bound of the target generalization error is first given based on the Chi-Square divergence, which estimates the difference of the joint distribution of common classes in the source and target domains. And then a UniDA model is built by minimizing this upper bound.  

The contributions of this paper can be summarized as:
\begin{itemize}
    \item An upper bound of the generalization error on the target domain is first proposed based on the Chi-Square divergence.
    \item  A novel UniDA model is built by minimizing the above upper bound of the generalization error.
    \item A progressive pseudo labeling method is designed to assign the pseudo labels to the unlabeled target samples, which helps to improve the classification performance of the unlabeled samples on the target domain. Based on this progressive pseudo labeling strategy, a novel UniDA algorithm called Joint Distribution Alignment for Universal Domain Adaptation (JAUA) is proposed.
    \item {\color{black}To verify the effectiveness of JAUA, JAUA is compared with twenty-four state-of-the-art methods on six public datasets. The results of the experiment show the superiority of JAUA in solving UniDA problems.}
\end{itemize}

The rest of this paper is organized as follows. Section II introduces the related work. Section III gives the whole model of our work. Section IV shows the complete experimental results and analysis. Section V gives conclusions and future work.
\section{Related work}
In this section, we introduce briefly the related work including universal domain adaptation, some UniDA methods based on progressive separation and theoretical analysis of open-set domain adaptation.
\subsection{Universal Domain Adaptation}
Most of the existing UniDA methods rely on threshold-based criteria to detect unknown-class samples in the target domain. To solve the UniDA problem, You \cite{you2019universal} proposed the Universal Adaptation Network (UAN), which assigns higher weights to samples from common classes by measuring sample-level transferability and employs a domain discriminator trained adversarially to align cross-domain features. Fu \cite{fu2020learning} introduced a Calibrated Multiple Uncertainty (CMU) metric, which estimates the probability of target samples belonging to common classes through hybrid uncertainty estimation. Li \cite{li2021domain} developed the Domain Consensus Clustering (DCC) method, leveraging domain consensus knowledge to cluster target-domain data while separating unknown classes from common classes. The DANCE framework \cite{saito2020universal} avoids relying solely on source supervision for discriminative representations; instead, it employs self-supervised learning to capture the target domain's clustering structure and learns feature representations through neighbor clustering. Lv \cite{lv2023lafea} introduced a GAN-style autoencoder to transform features into latent representations and performed adversarial domain adaptation in the latent space, thereby providing additional guidance to the backbone network to enhance feature compactness and domain discriminability. 

These methods have been widely adopted in UniDA research and provided valuable insights for addressing domain adaptation challenges in real-world applications. However, from the point of view of the principle, the key problem of UniDA is how to reduce the distribution difference between the source domain and the target domain and separate the unknown classes from the target domain.
\subsection{Methods Based on Progressive Separation}
For the UniDA problem, there exist unknown classes in the target domain. When the source domain is aligned with the target domain, if these unknown classes are not excluded, it will result in negative transferring due to the mismatch between the common classes and the unknown classes. Therefore, it is especially important to effectively separate the common class samples and the unknown class samples. To deal with this problem, Liu proposed Separate To Adaptive (STA) \cite{liu2019separate} which is one of the most representative methods. STA uses a progressive separation mechanism to separate the common class samples and the unknown class samples by two-stage method. In the coarse separation stage, STA uses a multi-binary classifier to estimate the similarity between the target domain samples and each source domain class. In the fine separation stage, it selects samples with very high and very low similarity as common and unknown class data, respectively to train binary classifiers. Based on the observation that due to the similarity of weights of the common class and unknown class samples in the early training stage, a large number of unknown class samples will be involved in the distributional alignment of the common class and such a negative effect may not be ameliorated in the later stage of training, Gao \cite{gao2020adversarial} proposed Threshold Domain Adversarial Network (ThDAN), which adaptively computes the threshold by quantifying the average transferability scores of the source domain samples and progressively selecting the common class to update the classifier during adversarial training. To eliminate false positive samples in the unknown classes, Dai \cite{dai2022gcl} proposed open-set domain adaptation GCL-OSDA for graph cooperative learning. In GCL-OSDA, an evidence network is utilized to output the confidence of the target domain samples based on the Dempster-Shafer (D-S) theory of evidence and the reliability of selecting the unknown samples is improved by asymptotically selecting the samples with a very high level of uncertainty. Du \cite{du2021self} proposed the Self Separation and Misseparation impact Minimization (SSMM) method, which progressively selects the target domain samples with low predictive entropy and high predictive entropy as the common class and unknown class, respectively, and maximizes the distribution difference between the common class and the unknown class. In this way, SSMM can minimize the effect of mis-separation of the target domain samples and reduce the algorithm's dependence on the similarity between the source and target domains.

Although the above algorithms are designed to solve the domain adaptive problem and can achieve good experimental results, they are not based on the target generalization error bounds and their theoretical interpretability is lacked.
\subsection{Generalization Error Bound}
In order to fill the theoretical gap of the adaptive problem in the open set domain, Fang\cite{fang2020open} proposed the generalization error bound of the target domain: 
\begin{equation}
\begin{aligned} \frac{R^t(h)}{1-\pi_{C+1}^t} \leq R^s(h) & +2 d_{\mathcal{H}}^{\ell}\left(P_{X^t \mid Y^s}, P_{X^s}\right) \\ & +\left(\frac{R_{u, C+1}^t(h)}{1-\pi_{C+1}^t}-R_{u, C+1}^s(h)\right)+\Lambda\label{eq1}\end{aligned}
\end{equation}
where $R^s (h)$ is the generalization error on the source domain, and $d_{\mathcal{H}}^{\ell}\left(P_{X^t \mid Y^s}, P_{X^s}\right)$ is the distribution difference between the common classes of the target domain and the source domain, which converts the MMD distance into a suitable regularization term for aligning marginal distributions and class-conditional distributions of the common classes among the source domain and target domain. $\triangle_0=\frac{R_{u, C+1}^t(h)}{1-\pi_{C+1}^t}-R_{u, C+1}^s(h)$ is the open-set difference which is considered to be closely related to the risk of the unknown classes. $\Lambda$ is the stream regularization term, which makes the similar samples closer and can be used to represent the geometrical relationship between the source and target domains. 

Although the above work addresses the open-set domain adaptation problem by investigating the generalized error bound theory, it ignores the negative transferable effects of separating all samples at once. At present, there is a lack of similar theoretical discussion for universal domain adaptation.
\section{Symbols and Problem Setting}
In this section, we first introduce briefly some mathematical symbols and notations. And then, we give problem setting. 
\subsection{Symbols and Notations}
We list all of the mathematical symbols and notations used in this study in Table \ref{tab:table1}.
\begin{table}[!t]
\caption{Symbols and notations\label{tab:table1}}
\centering
\begin{tabular}{|c||c|}
\hline
Symbols & Descriptions\\
\hline
$\mathcal{X}\in \mathcal{R}^D$ & Feature space\\
\hline
$C$ & The number of common classes\\
\hline
$\mathcal{Y}^{u_s}=\{y^{u_s}\}=\{C+1\}$ & Private label space\\
\hline
$\mathcal{Y}^{u_t}=\{y^{u_t}\}=\{C+2\}$ & Unknown label space\\
\hline
$\mathcal{Y}^{u_G}=\{y_1,y_2,...,y_c\}$ & Common label space\\
\hline
$\mathcal{Y}^s=\{y_1,y_2,...,y_c,y^{u_s}\}$ & Source label space\\
\hline
$\mathcal{Y}^t=\{y_1,y_2,...,y_c,y^{u_t}\}$ & Target label space\\
\hline
$P_{X^s Y^s}(x,y)$ & Source joint probability distribution\\
\hline
$P_{X^t Y^t}(x,y)$ & Target joint probability distribution\\
\hline
$P_{X^s}(x)$ & Source margin probability distribution\\
\hline
$P_{X^t}(x)$ & Target margin probability distribution\\
\hline
$m_s$ & The size of source domain\\
\hline
$m_t$ & The size of target domain\\
\hline
$D^s=\{x^s_i,y^s_i\}_{i=1}^{m_s}$ & Source domain\\
\hline
$D^t=\{x^t_i\}_{i=1}^{m_t}$ & Target domain\\
\hline
$D^s_i$ & The $i$-th class samples\\
& from source domain\\
\hline
$D^t_i$ & The $i$-th class samples\\
& from target domain\\
\hline
$D^t_u$ & The unknown class samples\\
& from target domain\\
\hline
$D^s_k$ & The common class samples\\
& from source domain\\
\hline
$D^t_k$ & The common class samples\\
& from target domain\\
\hline
$m_{tu}$ & The size of $D^t_u$\\
\hline
$m_{sk}$ & The size of $D^s_k$\\
\hline
$m_{tk}$ & The size of $D^t_k$\\
\hline
$h \in \mathcal{H}$ & Classification model\\
\hline
$R^s(h)$ & Source risk\\
\hline
$R^t(h)$ & Target risk\\
\hline
$R^{tu}(h)$ & Target unknown risk\\
\hline
$R^{tk}(h)$ & Target common risk\\
\hline
$R^{su}(h)$ & Source private risk\\
\hline
$R^{sk}(h)$ & source common risk\\
\hline
$R^{s}_{u_s}(h)$ & Loss of samples of the source domain\\& being classified as private class\\
\hline
$R^{t}_{u_t}(h)$ & Loss of samples of the target domain\\& being classified as unknown class\\
\hline
$R^{sk}_{u_t}(h)$ & Loss of common class samples \\& of the source domain being\\& misclassified as unknown class\\
\hline
$R^{tk}_{u_s}(h)$ & Loss of common class samples\\
& of the target domain being\\& misclassified as private class\\
\hline
$R^{tu}_{u_t}(h)$ & Loss of unknown class samples\\
& of the target domain being \\& classified as unknown class\\
\hline
$R^{tk}_{u_t}(h)$ & Loss of common class samples\\
& of the target domain being\\& misclassified as unknown class\\
\hline
$R^{su}_{u_s}(h)$ & Loss of private class samples\\
& of the source domain being\\& classified as private class\\
\hline
$R^{sk}_{u_s}(h)$ & Loss of common class samples\\
& of the source domain being\\& misclassified as private class\\
\hline
$R^{tk}_{k}(h)$ & Loss of common class samples\\
& of the target domain being\\& classified as common class\\
\hline
\end{tabular}
\end{table}
\subsection{Problem Setting}
\textbf{\textit{Problem}}: (Universal Domain Adaptation) Given source joint probability distribution $P_{X^s Y^s} (x,y)$ and target joint probability distribution $P_{X^t Y^t} (x,y)$, where $P_{X^t Y^t} (x,y)$ is different from $P_{X^s Y^s} (x,y)$. The label space of source domain is $\mathcal{Y}^s=\{y_1,y_2,\dots,y_c,y^{u_s }\}$. The label space of target domain is $\mathcal{Y}^t=\{y_1,y_2,\dots,y_c,y^{u_t}\}$ which is different from $\mathcal{Y}^s$ obviously. Given the source domain $D^s=\{x_i^s,y_i^s\}_{i=1}^{m_s}$ drawn from $P_{X^s Y^s}(x,y)$ i.i.d. and the target domain $D^t=\{x_i^t\}_{i=1}^{m_t}$ drawn from $P_{X^t} (x)$ i.i.d.. Our goal is to find an optimal classifier $\mathit{h}: \mathcal{X} \rightarrow \mathcal{Y}^t$ and make the expected risk $R^t(h)=E_{(x,y)\sim P_{X^t Y^t} (x,y)} [l(h(x),y)]=\int l(h(x),y) P_{X^t Y^t} (x,y)dxdy$ smallest, where $l(h(x),y)$ is a loss function to measure the accuracy of the classifier.

\section{Proposed Model And Algorithm}
In this section, we firstly give the upper bound of the generalization error on the target domain. Secondly, a novel mathematical model is built by minimizing the upper bound of the generalization error. Finally, a progressive pseudo labeling algorithm is designed.
\subsection{Upper Bound of Generalization Error}
The definitions of $R^t(h)$, $R^s(h)$, $R^t_{u_t}(h)$, $R^s_{u_s}(h)$, $R^{sk}_{u_t}(h)$ and $R^{tk}_{u_s}(h)$ are as follows:

\begin{equation}
R^s(h)=\int_{\mathcal{X}*\mathcal{Y}^s}l(h(x),y)P_{X^sY^s}(x,y)dxdy\label{eq2}
\end{equation}

\begin{equation}
R^t(h)=\int_{\mathcal{X}*\mathcal{Y}^t}l(h(x),y)P_{X^tY^t}(x,y)dxdy\label{eq3}
\end{equation}

\begin{equation}
\begin{aligned}
R^t_{u_t}(h)=&\int_{\mathcal{X}}l(h(x),y^{u_t})P_{X^t}(x)dx\\
=&\int_{\mathcal{X}*\mathcal{Y}^{u_t}}l(h(x),y^{u_t})P_{X^tY^t}(x,y)dxdy\\&+\int_{\mathcal{X}*\mathcal{Y}^{u_G}}l(h(x),y^{u_t})P_{X^tY^t}(x,y)dxdy\\
\triangleq& R^{tu}_{u_t}(h)+R^{tk}_{u_t}(h)\label{eq4}
\end{aligned}
\end{equation}

\begin{equation}
\begin{aligned}
R^{s}_{u_s}(h)=&\int_{\mathcal{X}}l(h(x),y^{u_s})P_{X^s}(x)dx\\
=&\int_{\mathcal{X}*\mathcal{Y}^{u_s}}l(h(x),y^{u_s})P_{X^sY^s}(x,y)dxdy\\&+\int_{\mathcal{X}*\mathcal{Y}^{u_G}}l(h(x),y^{u_s})P_{X^sY^s}(x,y)dxdy\\
\triangleq&R^{su}_{u_s}(h)+R^{sk}_{u_s}(h)\\\label{eq12}
\end{aligned}
\end{equation}

\begin{equation}
R^{sk}_{u_t}(h)=\int_{\mathcal{X}*\mathcal{Y}^{u_G}}l(h(x),y^{u_t})P_{X^sY^s}(x,y)dxdy\label{eq5}
\end{equation}

\begin{equation}
R^{tk}_{u_s}(h)=\int_{\mathcal{X}*\mathcal{Y}^{u_G}}l(h(x),y^{u_s})P_{X^tY^t}(x,y)dxdy\label{eq6}
\end{equation}

In this paper, we use Chi-Square divergence to calculate the difference between the joint distributions of the source domain and the target domain. Based on its following definition:
\begin{equation}
\begin{aligned}
&D_{CS}(P_{X^sY^s\sim\mathcal{X}*\mathcal{Y}^{u_G}}||P_{X^tY^t\sim\mathcal{X}*\mathcal{Y}^{u_G}})\\&=\int_{\mathcal{X}*\mathcal{Y}^{u_G}}P_{X^sY^s}(x,y)\frac{P_{X^sY^s}(x,y)}{P_{X^tY^t}(x,y)}dxdy-1\\&
\approx\frac{1}{m_{sk}} \sum_{i=1}^{m_{sk}} \frac{P_{X^sY^s}(x^s_i,y^s_i)}{P_{X^tY^t}(x^s_i,y^s_i)}-1\label{eq0}
\end{aligned}
\end{equation}
we can give the following \textbf{Theorem 1}.

\textbf{\textit{Theorem 1}}: Given a bounded loss function $\mathit{l}$. If there exist positive real numbers $M$, $N_s$ and $N_T$, which makes $|l(h(x),y)| \leq M$ hold for any $(x,y) \in \mathcal{X}*\mathcal{Y}^{u_G}$, $|l(h(x),y)| \leq N_s$ hold for any $(x,y) \in \mathcal{X}*\mathcal{Y}^{u_s}$, $|l(h(x),y)| \leq N_T$ hold for any $(x,y) \in \mathcal{X}*\mathcal{Y}^{u_t}$, then for any $\mathit{h} \in \mathcal{H}$, the following inequality holds:
\begin{equation}
\begin{aligned}
R^t(h) &\leq R^s(h)+\lambda_{t t} R_{u_t}^{tu}(h)+(\lambda_{t}+\lambda_s) R_{k}^{tk}(h) \\
& -\lambda_{gst}R_{u_t}^{sk}(h)-\lambda_{g s s} R_{u_s}^{sk}(h)+\left(3 M+N_T+N_S\right) \\
& * \sqrt{2 D_{CS}\left(P_{X^s Y^s \sim \mathcal{X}* \mathcal{Y}^{u_G}} \| P_{X^t Y^t \sim \mathcal{X}* \mathcal{Y}^{u_G}}\right)}+A+B\label{eq7}
\end{aligned}
\end{equation}
where $\lambda_{tt}$, $\lambda_{gst}$ and $\lambda_{gss}$ are hyperparameters, $A$ and $B$ are positive real numbers which satisfy $R^{tk}_{u_t}(h) \leq \frac{\lambda_t}{\lambda_{tt}}R^{tk}_k(h)+\frac{A}{\lambda_{tt}}$ and $R^{tk}_{u_s}(h) \leq \lambda_s R^{tk}_k(h)+B$, $\lambda_t$ and $\lambda_s$ are hyperparameters which bound $R^{tk}_{u_t}(h)$ and $R^{tk}_{u_s}(h)$. $R^{tk}_{k}(h)$ is the loss of common classes of the target domain being classified as common class and its definition is as follows:

\begin{equation}
R^{tk}_k(h)=\int_{\mathcal{X}*\mathcal{Y}^{u_G}}l(h(x),y)P_{X^tY^t}(x,y)dxdy\label{eq14}
\end{equation}

\color{black}The detailed proof of \textbf{Theorem 1} is given in the \textbf{Appendix A}.

\subsection{Proposed Model and Solution Method}
\textbf{Theorem 1} shows that the generalization error of the target domain is affected by the source domain risk $R^s (h)$, loss of unknown class samples of the target domain being classified as unknown class $R_{u_t}^{tu}(h)$, loss of common class samples of the target domain being classified as common class $R_{k}^{tk}(h)$, loss of common class samples of the source domain being misclassified as unknown class $R_{u_t}^{sk} (h)$, loss of common class samples of the source domain being misclassified as private class $R_{u_s}^{sk}(h)$ and the difference between the joint probability distributions of two domains. In order to minimize the generalization error of the target domain, we give the following optimization model based on inequality \eqref{eq7}

\begin{equation}
\begin{aligned}
\min_{h \in \mathcal{H}} &\hat{R}^s(h)+\lambda_{tt}\hat{R}^{tu}_{u_t}(h)+(\lambda_t+\lambda_s)\hat{R}^{tk}_k(h)\\
&-\lambda_{gst}\hat{R}^{sk}_{u_t}(h) -\lambda_{gss}\hat{R}^{sk}_{u_s}(h) +(3M+N_s+N_t)\\&*\sqrt {2 D_{CS}(P_{X^SY^S\sim\mathcal{X}*\mathcal{Y}^{u_G}}||P_{X^tY^t\sim\mathcal{X}*\mathcal{Y}^{u_G}})} \label{eq16}
\end{aligned}
\end{equation}

On the one hand, from the definition of Chi-Square divergence $D_{CS}(P_{X^SY^S\sim\mathcal{X}*\mathcal{Y}^{u_G}}||P_{X^tY^t\sim\mathcal{X}*\mathcal{Y}^{u_G}})$, we know that its estimation needs label information of the target domain samples which will be provided by $h$. On the other hand, in order to obtain the optimal $h$, we need to find the optimal $D_{CS}(P_{X^SY^S\sim\mathcal{X}*\mathcal{Y}^{u_G}}||P_{X^tY^t\sim\mathcal{X}*\mathcal{Y}^{u_G}})$. Based on the above considerations, we solve the optimization problem \eqref{eq16} based on alternative iteration scheme. First, we estimate $h$ based on Representer Theorem \cite{scholkopf2001generalized}. Then we estimate $D_{CS}(P_{X^SY^S\sim\mathcal{X}*\mathcal{Y}^{u_G}}||P_{X^tY^t\sim\mathcal{X}*\mathcal{Y}^{u_G}})$ based on $h$.

(1) Estimation of hypothesis $h$

Based on \eqref{eq16}, we estimate $h$ by solving the following problem:
\begin{equation}
\begin{aligned}
h^{*}=\arg\min_{h \in \mathcal{H}} & \hat{R}^s(h)+\lambda_{tt}\hat{R}^{tu}_{u_t}(h)+(\lambda_t+\lambda_s)\hat{R}^{tk}_k(h) \\
& -\lambda_{gst}\hat{R}^{sk}_{u_t}(h)-\lambda_{gss}\hat{R}^{sk}_{u_s}(h)+\eta \|h\|^2_{\mathcal{H}} \label{eq17}
\end{aligned}
\end{equation}

Using representer theorem, let $h$ be represented as:
\begin{equation}
h^{*}(x)=\sum_{i=1}^{m_s+m_t} \alpha_i k(x,x_i) \label{eq18}
\end{equation}
where $\alpha=(\alpha_1,\alpha_2,...,\alpha_{m_s+m_t})\in \mathcal{R}^{(m_s+m_t)*(C+2)}$ are coefficients and $(K)_{ij}=k(x_i,x_j)$, $K\in \mathcal{R}^{(m_s+m_t)*(m_s+m_t)}$ is a kernel matrix.

{\color{black}Inspired by \cite{chen2020subspace}, in this study, the following quadratic loss function $l$ is used:
\begin{equation}
l(h(x_i),y_i)=(y_i-h(x_i))^2 \label{eq19}
\end{equation}

Compared to the other loss, on the one hand, quadratic loss function satisfies the Lipschitz condition in any finite closed interval and this character provides the theoretical basis for proving effectiveness of the proposed progressive pseudo labeling method. On the other hand, it can be easily derived to yield an analytical solution for $h$.

Let $Y$, $Y_{u_t}$, $Y_k\in R^{(C+2)*(m_s+m_t)}$ be 0-1 matrices, $W$, $V_{tu}$, $V_k$, $V_{sk}\in R^{(m_s+m_t)*(m_s+m_t)}$ and $Y_{u_s}\in R^{(C+2)*(m_s+m_t)}$ be a diagonal matrices. Their specific definitions are as follows:
\begin{equation}
(Y)_{ij} = \begin{cases}
1, & x_j \in D_i^s \\
{0,} & {\text{otherwise.}}
\end{cases} \label{eq20}
\end{equation}

\begin{equation}
(Y_{u_t})_{ij} = \begin{cases}
1, & i=C+2 \\
{0,} & {\text{otherwise.}}
\end{cases} \label{eq21}
\end{equation}

\begin{equation}
(Y_{k})_{ij} = \begin{cases}
1, & x_j\in D^t_i \\
{0,} & {\text{otherwise.}}
\end{cases} \label{eq22}
\end{equation}

\begin{equation}
(W)_{ii} = \begin{cases}
\sqrt {\frac{1}{m_s}}, & x_i \in D^s \\
{0,} & {\text{otherwise.}}
\end{cases} \label{eq23}
\end{equation}

\begin{equation}
(V_{tu})_{ii} = \begin{cases}
\sqrt {\frac{1}{m_{tu}}}, & x_i \in D^t_u \\
{0,} & {\text{otherwise.}}
\end{cases} \label{eq24}
\end{equation}

\begin{equation}
(V_{k})_{ii} = \begin{cases}
\sqrt {\frac{1}{m_{tk}}}, & x_i \in D^t_k \\
{0,} & {\text{otherwise.}}
\end{cases} \label{eq25}
\end{equation}

\begin{equation}
(V_{sk})_{ii} = \begin{cases}
\sqrt {\frac{1}{m_{sk}}}, & x_i \in D^s_k \\
{0,} & {\text{otherwise.}}
\end{cases} \label{eq26}
\end{equation}

\begin{equation}
(Y_{u_s})_{ij} = \begin{cases}
1, & i=C+1 \\
{0,} & {\text{otherwise.}}
\end{cases} \label{eq27}
\end{equation}

Then, based on the law of large numbers and matrix representation, we can easily obtain the empirical estimations of $R^s(h)$, $R^{tu}_{u_t}(h)$, $R^{tk}_k(h)$, $R^{sk}_{u_t}(h)$ and $R^{sk}_{u_s}(h)$ as follows:

\begin{equation}
\hat{R}^s(h)=\|(Y-\alpha^\top K)W\|^2_F \label{eq28}
\end{equation}

\begin{equation}
\hat{R}^{tu}_{u_t}(h)=\|(Y_{u_t}-\alpha^\top K)V_{tu}\|^2_F \label{eq29}
\end{equation}

\begin{equation}
\hat{R}^{tk}_{k}(h)=\|(Y_{k}-\alpha^\top K)V_{k}\|^2_F \label{eq30}
\end{equation}

\begin{equation}
\hat{R}^{sk}_{u_t}(h)=\|(Y_{u_t}-\alpha^\top K)V_{sk}\|^2_F \label{eq31}
\end{equation}

\begin{equation}
\hat{R}^{sk}_{u_s}(h)=\|(Y_{u_s}-\alpha^\top K)V_{sk}\|^2_F \label{eq32}
\end{equation}}

Based on the above matrix representations, the optimization problem \eqref{eq16} can be rewritten as follows:
\begin{equation}
\begin{aligned}
&(\alpha^{*}) \\
=& \arg \min_{\alpha \in \mathcal{R}^{(m_s+m_t)*(C+2)}} \eta tr(\alpha^\top K \alpha)+\|(Y-\alpha^\top K)W\|^2_2 \\&+\lambda_{tt}\|(Y_{u_t}-\alpha^\top K)V_{tu}\|^2_2+ (\lambda_{s}+\lambda_{t})\|(Y_{k}-\alpha^\top K)V_{k}\|^2_2 \\
&- \lambda_{gst}\|(Y_{u_t}-\alpha^\top K)V_{sk}\|^2_2 -\lambda_{gss}\|(Y_{u_s}-\alpha^\top K)V_{sk}\|^2_2 \label{eq33}
\end{aligned}
\end{equation}

\textbf{\textit{Theorem 2}}: If and only if $\lambda_{gst}+\lambda_{gss}\leq \frac{m_{sk}}{m_s}$, the objective function in the optimization problem \eqref{eq33} has unique minimum.

{\color{black}The proof of \textbf{Theorem 2} is provided in the \textbf{Appendix B}.}

The closed solution of $\alpha$ is as follows:

\begin{equation}
\begin{aligned}
\alpha^{*}=&((W^2+\lambda_{tt}V^2_{tu}+(\lambda_t+\lambda_t)V^2_k-(\lambda_{gst}+\lambda_{gss})V^2_{sk})K\\
&+\eta I)^{-1}(W^2Y^\top+\lambda_{tt}V^2_{tu}Y^2_{u_t}+(\lambda_t+\lambda_s)V^2_kY^2_k\\
&-\lambda_{gst}V^2_{sk}Y^2_{u_t}-\lambda_{gss}V^2_{sk}Y^2_{u_s})\label{eq34}
\end{aligned}    
\end{equation}

{\color{black}(2) Estimation of $D_{CS}(P_{X^SY^S\sim\mathcal{X}*\mathcal{Y}^{u_G}}||P_{X^tY^t\sim\mathcal{X}*\mathcal{Y}^{u_G}})$

After obtaining $h$, we use it to classify the samples of the target domain and estimate \eqref{eq0}. Inspired by \cite{ren2022buresnet,chen2024multi,chen2020domain,wen2024maximum,jin2020joint,chen2020semi,chen2022domain,chen2023domain}, we use kernel function $r(x,y;\rho)$ to approximate $\frac{P_{X^sY^s}(x^s_i,y^s_i)}{P_{X^tY^t}(x^s_i,y^s_i)}$.} Let $r(x,y;\rho)=\frac{P_{X^s Y^s}(x_i^s,y_i^s)}{P_{X^t Y^t}(x_i^s,y_i^s)}$ be expressed as
\begin{equation}
r(x,y;\rho)=\sum_{i=1}^{m_{tk}+m_{sk}} \rho_i\Phi_i(x,y)\label{eq35}
\end{equation}
where $\rho_i$ is the weight parameter. $\Phi_i(x,y)=k(x,x_i)l(y,y_i)$ is the kernel function, where $k(x,x_i)=exp(-\frac{||x-x_i||^2}{\sigma_x})$, $\sigma_x$ is the median inter-sample distance of $m_{tk}+m_{sk}$ samples. $l(y,y_i)$ is the delta kernel function, $l(y,y_i)=1$ if and only if $y=y_i$, otherwise $l(y,y_i)=0$.

From
\begin{equation}
\begin{aligned}
&\mathcal{J}(\rho) \\
= & \frac{1}{2}\int_{\mathcal{X}*\mathcal{Y}^{u_G}}[r(x,y;\rho)-\frac{P_{X^s Y^s}(x,y)}{P_{X^t Y^t}(x,y)}]^2P_{X^t Y^t}(x,y)dxdy \\
= & \frac{1}{2}\int_{\mathcal{X}*\mathcal{Y}^{u_G}}r^2(x,y;\rho)P_{X^t Y^t}(x,y)dxdy \\
&-\int_{\mathcal{X}*\mathcal{Y}^{u_G}}r(x,y;\rho)P_{X^s Y^s}(x,y)dxdy+C \\
= & \frac{1}{2}\sum_{i,j=1}^{m_{tk}+m_{sk}}\rho_i\rho_j\int_{\mathcal{X}*\mathcal{Y}^{u_G}}\Phi_i(x,y)\Phi_j(x,y)P_{X^t Y^t}(x,y)dxdy \\
&-\sum_{i=1}^{m_{tk}+m_{sk}}\rho_i\int_{\mathcal{X}*\mathcal{Y}^{u_G}}\Phi_i(x,y)P_{X^s Y^s}(x,y)dxdy+C \\
= & \frac{1}{2}\rho^\top H\rho-b^\top \rho+C\label{eq36}
\end{aligned}
\end{equation}
where $H\in R^{(m_{tk}+m_{sk})*(m_{tk}+m_{sk})}$, $b\in R^{m_{tk}+m_{sk}}$, and
\begin{equation}
(H)_{ij}=\int_{\mathcal{X}*\mathcal{Y}^{u_G}}\Phi_i(x,y)\Phi_j(x,y)P_{X^t Y^t}(x,y)dxdy\label{eq37}
\end{equation}
\begin{equation}
(b)_{i}=\int_{\mathcal{X}*\mathcal{Y}^{u_G}}\Phi_i(x,y)P_{X^s Y^s}(x,y)dxdy\label{eq38}
\end{equation}
We can estimate $\rho$ by solving the following optimization problem:
\begin{equation}
\rho^* = \arg\min_{\rho \in \mathbb{R}^{m_{tk} + m_{sk}}} \frac{1}{2} \rho^\top \hat{H} \rho - \hat{b}^\top \rho + \frac{\lambda}{2} \rho^\top \rho = (\hat{H} + \lambda I)^{-1} \hat{b} \label{eq39}
\end{equation}
where $\frac{\lambda}{2} \rho^\top \rho$ is the penalty item avoiding overfitting,
\begin{equation}
(\hat{H})_{ij}=\frac{1}{m_{tk}}\sum_{k=1}^{m_{tk}}\Phi_i(x^t_k,y^t_k)\Phi_j(x^t_k,y^t_k)\label{eq40}
\end{equation}
\begin{equation}
(\hat{b})_{i}=\frac{1}{m_{sk}}\sum_{k=1}^{m_{sk}}\Phi_i(x^s_k,y^s_k)\label{eq41}
\end{equation}

Hence,
\begin{equation}
\begin{aligned}
&D_{CS} (P_{X^s Y^s \sim \mathcal{X}* \mathcal{Y}^{u_G}} \| P_{X^t Y^t \sim \mathcal{X}* \mathcal{Y}^{u_G}}) \\
\approx & \frac{1}{m_{sk}}\sum_{i=1}^{m_{sk}}max\{r(x^s_i,y^s_i;(\hat{H}+\lambda I)^{-1} \hat{b}),\varepsilon \}-1\label{eq42}
\end{aligned}
\end{equation}

{\color{black}From \eqref{eq42}, we know that the sample features of the source domain and target domain have important effects on $r(x,y;\rho)$. To obtain the better $r(x,y;\rho)$, we first use a pair of projection matrices $W_s$ and $W_t$ to project the source samples and the target samples into the high-dimensional space, respectively. Then, we minimize $D_{CS} (P_{X^s Y^s \sim \mathcal{X}* \mathcal{Y}^{u_G}} \| P_{X^t Y^t \sim \mathcal{X}* \mathcal{Y}^{u_G}})$ to align the joint distribution of the source and target domains. Obviously, the optimization problem \eqref{eq42} can be rewritten as follows:}
\begin{equation}
\begin{aligned}
&\min_{W_s, W_t} f(W_s, W_t) \\
&= \min_{W_s, W_t} \frac{1}{m_{sk}} \sum_{i=1}^{m_{sk}} \max\left\{ r\left(W_s^\top x_i^s, y_i^s; (\hat{H} + \lambda I)^{-1} \hat{b}\right), \varepsilon \right\} \\
&s.t. \ W_s^\top W_s = I, \ W_t^\top W_t = I \label{eq43}
\end{aligned}
\end{equation}
where $W_s,W_t\in \mathcal{R}^{D*d},d \leq D$ and
\begin{equation}
\hat{H}=\frac{1}{m_{sk}}\sum_{k=1}^{m_{sk}}\Phi_i(W_s^\top x^s_k,y^s_k)\Phi_j(W_s^\top x^s_k,y^s_k) \label{eq44}
\end{equation}
\begin{equation}
\hat{b}=\frac{1}{m_{tk}}\sum_{k=1}^{m_{tk}}\Phi_i(W_t^\top x^t_k,y^t_k) \label{eq45}
\end{equation}

In this study, we use Steepest Descent (SD) \cite{meza2010steepest} to solve the optimization problem \eqref{eq43}. After obtaining $r(x,y;\rho)$, we fix it to calculate $\alpha$ by \eqref{eq34}.

\subsection{Progressive Pseudo Labeling Method}
{\color{black}In order to solve the optimization problem \eqref{eq16}, the pseudo labels of the target samples need to be provided. In the previous researches \cite{liu2019separate,gao2020adversarial,dai2022gcl,du2021self,luo2023source,luo2020progressive}, progressive labeling methods have been proposed to assign high confidential pseudo labels into the target samples. Specifically, in the initial stage, they first use the source samples to train a classifier and then assign the target samples into the specific classes. In the subsequent iteration stages, based on the confidential degrees of the pseudo labels, they build the optimization model by progressively selecting a part of the target samples. Generally speaking, since there are no unknown class labels in the label space of the source domain, the classifier cannot assign unknown class labels to the target samples directly.
Previous methods have largely relied on tools such as entropy and pre-classifiers as criteria for sample separation, but these approaches may overlook the spatial structure of samples, resulting in outcomes that lack interpretability.} Notice that samples that have the same label tend to be more similar, so the labels of the samples can be classified based on the similarity among the samples. Specifically, if some target samples are significantly different from all the source samples, then the classifier can assign them into unknown class. In this study, {\color{black}SVM \cite{xue2009svm} is used as classifier. Although SVM cannot assign unknown class labels to target samples directly, SVM can provide the confidence $E(x^t)=\{ E_1(x^t), E_2(x^t), ...,E_C(x^t)\}$ where $E_i$ represents the confidence score which represents the signed distance of the target sample $x^t$ to the $i$-th hyperplane.} Then the classes of the target samples $\{x^t_j\}^{m_t}_{j=1}$ can be determined by
\begin{equation}
\hat{y}_{j}^{t} = \begin{cases}
SVM(x^t_j), & max E(x^t_j) \geq \tau \\
C+2, & max E(x^t_j) < \tau
\end{cases} \label{eq46}
\end{equation}
where $\tau$ is the confidence threshold.

In practice, if SVM is used only once, the prediction will be inaccurate because there will still be some common class samples predicted as unknown class. To deal with this problem, we design a progressive labeling method. Specifically, let the total iteration number and the current iteration number be $T$ and $t$, respectively. In the $t$-th iteration, the target domain samples whose pseudo-labels are common classes will be selected at the ratio $t/T$. The reason why only the common class samples are selected is that the purpose of optimizing $D_{CS} (P_{X^s Y^s \sim \mathcal{X}*\mathcal{Y}^{u_G} } ||P_{X^t Y^t\sim \mathcal{X}*\mathcal{Y}^{u_G }} )$ is to align the joint probability distributions of common classes between the source and target domains. In the $t$-th iteration, $M_{tk}$ samples in each class are selected based on the ascending order of confidence as the common class samples where $M_{tk}$ is as follows:
\begin{equation}
M_{tk}=ceil(max(m_1,m_2,...,m_C)(\frac{t}{T}))\label{eq47}
\end{equation}

In particular, if $m_c \leq M_{tk}$, there will be only $m_c$ samples are selected. The confidence sequence of the selected samples in the $c$-th class can be expressed as:
\begin{equation}
\begin{aligned}
l_c =& \begin{cases}
E(x^t_{c,1}),E(x^t_{c,2}),...,E(x^t_{c,m_c}), & {\text{if}}\ m_c \leq M_{tk} \\
E(x^t_{c,1}),E(x^t_{c,2}),...,E(x^t_{c,M_{tk}}), & {\text{if}}\ m_c > M_{tk}
\end{cases} \\
& s.t. \ E(x_{c,i}^t) > E(x_{c,j}^t) \ {\text{if}}\ i < j \label{eq48}
\end{aligned}
\end{equation}

\begin{figure*}[!t]
\centering
\includegraphics[width=6in]{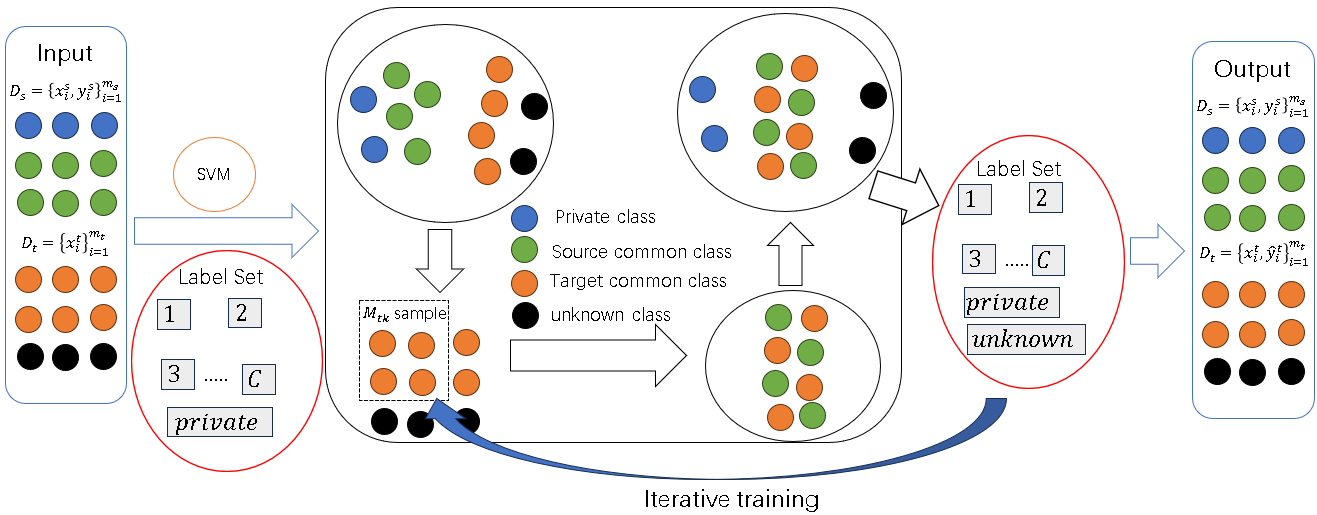}
\caption{{\color{black}The framework of JAUA. The input dataset is first classified into $1,2,\ldots,C+1$ classes using SVM. Through progressive pseudo-labeling, the dataset's classification is progressively updated. Target domain samples with confidence $E$ below $\tau$ are classified as $C+2$, and the label set is updated. This iterative training process is repeated until the final target domain classification results are obtained.}}
\label{fig3}
\end{figure*}

In this way, some common class samples are selected at each iteration to participate in the alignment process of the joint distributions between the source and target domains. {\color{black}In order to make the reader understand our idea clearly, the framework of JAUA is shown in Fig. \ref{fig3}.}

In the previous research, Chen \cite{chen2020self} pointed out that during the progressive pseudo-labeling process, errors in the initial pseudo-labels may propagate iteratively throughout training, potentially impacting algorithmic performance. For JAUA, the initial pseudo labels are usually also unreliable. However, they can be iteratively updated by the progressive pseudo labeling in the training phase to enhance their reliability. The theoretical analysis and discussions on error propagation issue from pseudo-labels to distribution alignment are provided in Supplementary Material. The related results show that the proposed progressive pseudo labeling method can adaptively reduce error propagation and align the joint distributions between two domains.

\subsection{Proposed Algorithm}
Based on the solution procedure of the optimization problem \eqref{eq16}, we first use SVM to obtain the pseudo labels of the target domain samples. And then Principal Component Analysis (PCA) \cite{abdi2010principal} is used to reduce the dimension of $D_s$ and $D_t$ from $D$ to $d$ and obtain initial $W_s$ and $W_t$. According to \eqref{eq33}, we can use $W_s$ and $W_t$ to get the updated $\alpha$, the updated classification model $h$ can be obtained based on \eqref{eq17}. In this way, the updated pseudo-label $\hat{y}^t$ can be obtained. In the subsequent iterations, the new common class samples can be selected by the new pseudo-labels to optimize \eqref{eq43} and get the new $W_s$ and $W_t$.

Based on the above analysis, the whole procedure of JAUA is summarized as \ref{algo1}.
\begin{algorithm}
\caption{JAUA}\label{algo1}
\begin{algorithmic}[1]
\REQUIRE {Source labeled dataset: $D_s=\{x^s_i,y^s_i\}^{m_x}_{i=1}$; Target unlabeled dataset: $D_t=\{x^t_i\}^{m_t}_{i=1}$; $m=m_s+m_t$; $m_k=m_{sk}+m_{tk}$; Hyperparameters $\lambda_{tt}$, $\lambda_s$, $\lambda_t$, $\lambda_{gss}$. $\lambda_{gst}$; Regular norm hyperparameters: $\lambda$, $\eta$; threshold of confidence: $\tau$; PCA dimension $d$; Total numbers of iterations: $T$; Iterations of aligning joint distribution: $S$; Bound of relative loss $\mathcal{E}$.
}
\ENSURE {Target domain predict labels: $\{\hat{y}_i^t\}^{m_t}_{i=1}$.
}
\STATE Init pseudo labels of target domain samples $\{\hat{y}^t_i\}^{m_t}_{i=1}$ by $SVM(D_s,D_t)$; 
\STATE Divide the dataset $D_t$ by \eqref{eq46} to obtain $D_{tk}$ and $D_{tu}$;
\STATE Init $W_s$, $W_t$ by $PCA((X_s,X_t),d)$;
\STATE Set the current total iteration number $t$ to 1;
\STATE Compute $M_{tk}$ by \eqref{eq47};
\STATE Obtain the confidence sequence $l_c$ by \eqref{eq48};
\STATE Set the current total loss $\varepsilon_0$ to 0;
\WHILE {$t<T$ and $\varepsilon<\mathcal{E}$}
        \STATE Set the iteration number of aligning joint distribution $s$ to 1;
        \WHILE {$s<S$}
        \STATE Use $l_c$ to optimize \eqref{eq43} by SD algorithm to update $W_s$, $W_t$;
        \STATE $s\leftarrow s+1$;
        \ENDWHILE
        \STATE Map samples $X_s$, $X_t$ to high dimensional samples $W_sX_s$,$W_tX_t$;
        \STATE Use $W_sX_s$,$W_tX_t$ to calculate $\alpha^*$ by \eqref{eq33};
        \STATE Compute the total loss $\varepsilon_t$ by \eqref{eq42};
        \STATE Update $\{\hat{y}^t_i\}^{m_t}_{i=1}\leftarrow\alpha^*K$;
        \STATE Update $M_{tk}$ by \eqref{eq47};
        \STATE Update the confidence sequence $l_c$ by \eqref{eq48};
        \STATE $t\leftarrow t+1, \varepsilon\leftarrow (\varepsilon_t-\varepsilon_{t-1})/\varepsilon_t$;
        \ENDWHILE
\STATE {return $\{\hat{y}^t_i\}^{m_t}_{i=1}$};
\end{algorithmic}
\end{algorithm}

{\color{black}Since JAUA needs to compute kernel matrix and matrix inversion in each iteration, the total time complexity and space complexity of JAUA are $O(T(m^3+Sm^3_k))$ and $O(m^2)$ respectively. For large scale data set, we choose to use Random Fourier Features \cite{rahimi2007random} and Conjugate Gradient \cite{hestenes1952methods} to compute kernel matrix and matrix inversion respectively. The total time complexity and space complexity can be reduced to $O(T(mdp+mp^2))$ and $O(mp)$, respectively. $p<<m$ represents the random feature dimension.}
\section{Experiment}
In this section, we conduct the experiments to verify the effectiveness of JAUA. JAUA algorithm is written in Python 3.9. The experiments are conducted on a computer with 13th Gen Intel(R) Core(TM) i7-13650HX.
\begin{table}[!t]
\caption{Dataset description\label{tab:table2}}
\centering
\begin{tabular}{|c|c|c|c|}
\hline
datasets & Domain & Sample number & Category \\
\hline
\multirow{3}{*}{Office-31} & Amazon & 2,817 & 31 \\
\cline{2-4} 
 & Dslr & 795 & 31 \\
\cline{2-4} 
 & Webcam & 498 & 31 \\
\hline
\multirow{4}{*}{Office-Home} & Art & 2,427 & 65 \\
\cline{2-4} 
 & Clipart & 4,365 & 65 \\
\cline{2-4} 
 & Product & 4,439 & 65 \\
\cline{2-4} 
 & RealWorld & 4,357 & 65 \\
\hline
\multirow{2}{*}{Visda} & Synthetic & 150,000 & 12 \\
\cline{2-4} 
 & RealWorld & 50,000 & 12 \\
\hline
\multirow{3}{*}{Domainnet} & Painting & 151,518 & 345 \\
\cline{2-4} 
 & Real & 350,654 & 345 \\
\cline{2-4} 
 & Sketch & 140,772 & 345 \\
\hline
\multirow{3}{*}{ImageCLEF} & Caltech & 600 & 12 \\
\cline{2-4} 
 & ImageNet & 600 & 12 \\
\cline{2-4} 
 & PASCAL VOC & 600 & 12 \\
\hline
\multirow{3}{*}{PACS} & Cartoon & 4,688 & 7 \\
\cline{2-4} 
 & Photo & 3,340 & 7 \\
\cline{2-4} 
 & Sketch & 7,858 & 7 \\
\hline
\end{tabular}
\end{table}

\subsection{Datasets}
Six public image datasets including Office-31\cite{saenko2010adapting}, Office-Home\cite{venkateswara2017deep}, Visda\cite{peng2017visda}, Domainnet\cite{peng2019moment}, ImageCLEF\cite{dos2025mlnet} and PACS\cite{li2017deeper}, are used for the experimental study and their overviews are provided in Table \ref{tab:table2}. The relevant settings for the datasets have been kept consistent with those of the baseline algorithms.
\begin{table*}[!t]
\caption{The parameter settings in JAUA\label{tab:table3}}
\centering
\begin{tabular}{|c|c|c|c|c|c|c|}
\hline
Parameters & Office-31 & Office-Home & Visda & Domainnet & ImageCLEF & PACS\\
\hline
$\lambda_{tt}$ & 1 & 1 & 1 & 1 & 1 & 1 \\
\hline
$\lambda_s$ & 0.1 & 0.1 & 0.07 & 0.1 & 0.07 & 0.1 \\
\hline
$\lambda_t$ & 0.1 & 0.1 & 0.07 & 0.1 & 0.07 & 0.1 \\
\hline
$\lambda_{gss}$ & 0.1 & 0.01 & 0.4 & 0.3 & 0.4 & 0.3 \\
\hline
$\lambda_{gst}$ & 0.3 & 0.01 & 0.2 & 0.4 & 0.2 & 0.3  \\
\hline
$\lambda$ & 0.0001 & 0.0001 & 0.0001 & 0.0001 & 0.0001 & 0.0001 \\
\hline
$\eta$ & 0.005 & 0.00015 & 0.005 & 0.00005 & 0.005 & 0.005 \\
\hline
$\tau$ & -0.1 & -0.5 & -0.1 & -0.5 & 0.5 & 0.1 \\
\hline
$S$ & 50 & 30 & 50 & 30 & 50 & 30 \\
\hline
$d$ & 100 & 300 & 100 & 300 & 100 & 300\\
\hline
$T$ & 20 & 20 & 20 & 20 & 20 & 20 \\
\hline
$\varepsilon$ & 1\% & 1\% & 1\% & 1\% & 1\% & 1\% \\
\hline
\end{tabular}
\end{table*}

\subsection{Baseline Algorithms}
We use four categories of baseline algorithms for comparison: 1) algorithms based on adversarial learning and feature alignment:UAN\cite{you2019universal}, CMU\cite{fu2020learning}, DANCE\cite{saito2020universal}, DCC\cite{li2021domain}, OSBP\cite{saito2018open}, IWAN\cite{zhang2018importance}, LaFea\cite{lv2023lafea}, Tanet\cite{wu2023tanet}, GUDA\cite{su2024generalized}, Mahalanobis\cite{fan2025mahalanobis}, BSP-WSA\cite{wang2024batch} and GATE \cite{chen2022geometric}; 2) algorithms based on classifier design and threshold strategies: OVANet\cite{saito2021ovanet}, CPR \cite{hur2023learning}, KLS\cite{wang2024exploiting}, PCL\cite{shan2023prediction}, STUN \cite{jain2024stochastic} and UACP \cite{wang2024towards}; 3) algorithms based on mutual learning and graph structures: MLNET\cite{lu2024mlnet}, E-MLNET \cite{dos2025mlnet}, UniDA-CGL\cite{fan2024curriculum} and LIWUDA\cite{zhu2024versatile}; and 4) algorithms based on contrastive learning and clustering: CAN\cite{yu2024novel} and MOEO\cite{ai2024maximum}.

\begin{table}[!t]
\caption{Comparison of the results on Visda dataset(Unit:\%)\label{tab:table4}}
\centering
\begin{tabular}{|c|c|c|c|}
\hline
Methods & ACC & UNK & HOS \\
\hline
UAN & 60.8 & 20.4 & 30.5 \\
\hline
CMU & 61.4 & 24.1 & 34.6 \\
\hline
DANCE & 69.2 & 31.0 & 42.8 \\
\hline
DCC & 64.2 & 32.3 & 43.0 \\
\hline
OSBP & 30.3 & \underline{49.9} & 37.7 \\
\hline
IWAN & 58.7 & 18.0 & 27.6 \\
\hline
OVANet & \underline{75.3} & 41.0 & 53.1 \\
\hline
LIWUDA & 64.2 & 27.6 & 38.6 \\
\hline
CPR & - & - & 58.2 \\
\hline
LaFea & - & - & 58.9 \\
\hline
Tanet & - & - & 60.1 \\
\hline
UniDA-CGL & - & - & 57.9 \\
\hline
Mahalanobis & - & - & 56.8 \\
\hline
BSP-WSA & - & - & 57.1 \\
\hline
CAN & - & - & \underline{61.6} \\
\hline
MOEO & - & - & 60.7 \\
\hline
GATE & - & - & 56.4 \\
\hline
KLS & - & - & 54.7 \\
\hline
UACP & - & - & 60.1 \\
\hline
JAUA & \textbf{75.5} & \textbf{52.6} & \textbf{62.0} \\
\hline
\end{tabular}
\end{table}

\subsection{Experimental evaluation metrics}
The performances of all algorithms are evaluated by the average accuracy of common class ACC, the accuracy of unknown class UNK and H-score (HOS). They are defined as:
\begin{equation}
ACC=\frac{1}{C}\sum_{i=1}^C \frac{|\{x|x \in D^i_t ~and ~ h(x)=y_i\}|}{|\{x|x\in D^i_t\}|}\label{eq49}
\end{equation}
\begin{equation}
UNK=\frac{|\{x|x \in D^{C+2}_t ~and ~ h(x)=y_{C+2}\}|}{|\{x|x\in D^{C+2}_t\}|}\label{eq50}
\end{equation}
\begin{equation}
HOS=\frac{2*ACC*UNK}{ACC+UNK}\label{eq51}
\end{equation}
where $h(x)$ represents the predict label of the sample $x$. $| \cdot |$ denotes the number of elements in the set. From the definitions of these metrics, we know that their ranges are [0,1]. Based on these definitions, we know that the closer the metrics are to 1, the better the algorithm performs.

\begin{table}[!t]
\caption{Comparison of the results on Office-31 dataset(Unit:\%)\label{tab:table5}}
\centering
\begin{tabular}{|c|c|c|c|}
\hline
Methods & ACC & UNK & HOS \\
\hline
UAN & 89.3 & 49.3 & 63.5 \\
\hline
CMU & 91.1 & 61.1 & 73.1 \\
\hline
DANCE & 93.0 & 78.2 & 84.7 \\
\hline
DCC & \underline{93.2} & 70.7 & 80.2 \\
\hline
IWAN & 86.7 & 36.7 & 51.6 \\
\hline
OVANet & 84.2 & \underline{88.0} & 85.9 \\
\hline
LIWUDA & 91.6 & 74.5 & 82.1 \\
\hline
PCL & 76.5 & 83.3 & 78.4 \\
\hline
CPR & - & - & 88.8 \\
\hline
LaFea & - & - & 90.5 \\
\hline
STUN & - & - & 89.8 \\
\hline
KLS & - & - & 88.5 \\
\hline
UACP & - & - & \underline{91.0} \\
\hline
JAUA & \textbf{93.4} & \textbf{89.0} & \textbf{91.1} \\
\hline
\end{tabular}
\end{table}

\subsection{Implementation Details}
In this study, we use the baseline network ResNet50 to extract the data features and the initial dimension of the feature is 2048. The parameter settings in JAUA are listed in Table \ref{tab:table3}. Except for $T$ and $\mathcal{E}$, the optimal values of the other parameters are obtained by the grid search.
\subsection{Experimental Results and Analysis}

Due to the page limitation, the average results of ACC, UNK and HOS on the datasets Visda, Office-31, Office-Home, Domainnet, ImageCLEF and PACS are listed in Table \ref{tab:table4}, Table \ref{tab:table5}, Table \ref{tab:table6}, Table \ref{tab:table7}, Table \ref{tab:table8} and Table \ref{tab:table9}, respectively. Except for Visda, the detailed results on each domain adaptation task are provided in Table S-3, Table S-4, Table S-5, Table S-6 and Table S-7, respectively, which are placed in Supplementary Material. In these Tables, the bold types and the underlined types represent the best results and the second-best results of different domain adaptation tasks among the comparison algorithms, respectively. “-” indicates the experimental results which are not provided by the original algorithms. In order to show JAUA’s scalability, the runtime of JAUA on six datsets are reported in Table S-2 of Supplementary Material.

\begin{table}[!t]
\caption{Comparison of the results on Office-Home
dataset(Unit:\%)\label{tab:table6}}
\centering
\begin{tabular}{|c|c|c|c|}
\hline
Methods & ACC & UNK & HOS \\
\hline
UAN & 77.0 & 45.2 & 56.6 \\
\hline
CMU & 77.2 & 52.0 & 61.6 \\
\hline
DANCE & 80.3 & 54.0 & 63.9 \\
\hline
DCC & 78.9 & \underline{68.4} & 72.0 \\
\hline
OSBP & 64.6 & 34.5 & 44.5 \\
\hline
IWAN & 73.3 & 33.0 & 45.3 \\
\hline
OVANet & - & - & 71.2 \\
\hline
LIWUDA & - & - & 62.7 \\
\hline
PCL & - & - & 70.3 \\
\hline
CPR & - & - & 72.3 \\
\hline
LaFea & - & - & 73.3 \\
\hline
STUN & - & - & 73.8 \\
\hline
KLS & - & - & 74.6 \\
\hline
UACP & - & - & \underline{74.7} \\
\hline
GUDA & 70.4 & - & - \\
\hline
JAUA & \textbf{80.4} & \textbf{71.4} & \textbf{75.4} \\
\hline
\end{tabular}
\end{table}

From \Cref{tab:table4,tab:table5,tab:table6,tab:table7,tab:table8,tab:table9}, we can find that compared with the baseline algorithms, the average ACC, UNK and HOS of JAUA on all of the datasets are optimal. The main reason why JAUA obtains the better results is as follows. {\color{black}On the one hand, JAUA aligns the joint probability distributions between the common classes of the source and target domains by minimizing Chi-Square divergence and learns more excellent feature representation. On the other hand, the baseline algorithms usually identify the unknown class samples by aligning the marginal distributions or the conditional distributions of the common classes.

From Table S-3 of Supplementary Material, we can find that on the tasks of D$\rightarrow$A and D$\rightarrow$W of the dataset Office-31 , in terms of UNK, JAUA shows more significant improvements. The main reason is that D only includes low-noise samples. When taking it as the source domain, JAUA can easily identify the unknown class samples.

From Table S-4 of Supplementary Material, we can find that on the tasks of Ar$\rightarrow$Cl, Ar$\rightarrow$Pr and Ar$\rightarrow$Rw of the dataset Office-Home , in terms of UNK, JAUA also shows more significant improvements. The main reason is that the samples in Ar primarily consist of sketches. When taking it as the source domain, JAUA can more effectively identify latent features of the unknown class samples. On the tasks of Ar$\rightarrow$Pr, Cl$\rightarrow$Pr and Rw$\rightarrow$Pr of the dataset Office-Home , in terms of ACC, JAUA obtains the best results. This may be attributed to Pr only containing background-free images with minimal redundant information. When taking it as the target domain, JAUA can easily recognize the common classes. When taking Rw as the target domain, all metrics of JAUA are improved. This indicates JAUA's strong classification capability for real-world images, reflecting its superior practical application potential in real-world scenarios. }

\begin{figure}[!t]
\centering
\includegraphics[width=3in]{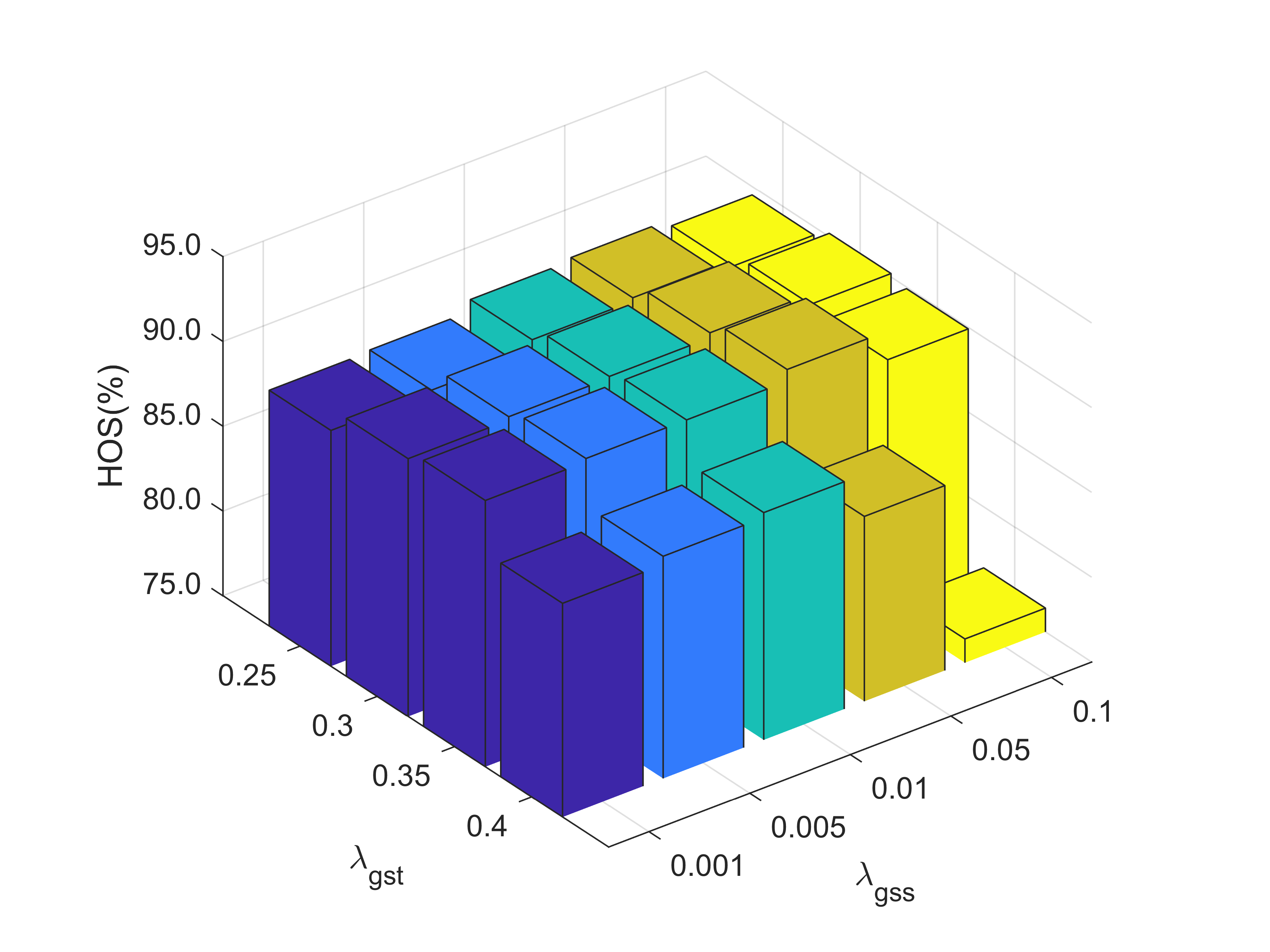}
\caption{Influence of the parameters of JAUA for task: D$\rightarrow$A on Office-31}
\label{fig4}
\end{figure}

\subsection{Parameter Sensitivity Analysis}
JAUA includes five hyperparameters $\lambda_{tt}$, $\lambda_s$, $\lambda_t$, $\lambda_{gss}$ and $\lambda_{gst}$. In this subsection, we explore the effects of these hyperparameters on JAUA by fixing three parameters and adjusting the others. The experiments are conducted on the dataset Office-31.

\begin{table}[!t]
\caption{Comparison of the results on Domainnet
dataset(Unit:\%)\label{tab:table7}}
\centering
\begin{tabular}{|c|c|}
\hline
Methods & HOS \\
\hline
UAN & 41.0 \\
\hline
CMU & 48.3 \\
\hline
DANCE & 33.5 \\
\hline
DCC & 49.2 \\
\hline
OSBP & 32.0 \\
\hline
IWAN & 32.8 \\
\hline
OVANet & 50.7 \\
\hline
LIWUDA & 48.6 \\
\hline
PCL & 41.3 \\
\hline
LaFea & 51.9 \\
\hline
Tanet & \underline{57.4} \\
\hline
CAN & 52.1 \\
\hline
MOEO & 52.6 \\
\hline
GATE & 52.1 \\
\hline
KLS & 51.8\\
\hline
UACP & 50.3 \\
\hline
JAUA & \textbf{58.1} \\
\hline
\end{tabular}
\end{table}

\begin{table}[!t]
\caption{Comparison of the results on ImageCLEF
dataset(Unit:\%)\label{tab:table8}}
\centering
\begin{tabular}{|c|c|}
\hline
Methods & HOS \\
\hline
OVANet & 74.3 \\
\hline
MLNET & 83.3\\
\hline
E-MLNET & \underline{85.4} \\
\hline
JAUA & \textbf{89.4} \\
\hline
\end{tabular}
\end{table}

\begin{table}[!t]
\caption{Comparison of the results on PACS
dataset(Unit:\%)\label{tab:table9}}
\centering
\begin{tabular}{|c|c|}
\hline
Methods & ACC \\
\hline
UAN & 30.7 \\
\hline
CMU & 21.0 \\
\hline
DANCE & 17.6 \\
\hline
OVANet & \underline{58.5} \\
\hline
GUDA & 58.2\\
\hline
JAUA & \textbf{64.2} \\
\hline
\end{tabular}
\end{table}

Due to space limitation, we only give the results for the task D$\rightarrow$A. We fix $\lambda_{tt}=1$, $\lambda_s=0.7$ and $\lambda_t=0.7$. The changes of HOS with respect to $\lambda_{gss}$ and $\lambda_{gst}$ are illustrated in Fig. \ref{fig4}. Fig. \ref{fig4} shows a three-dimensional histogram of HOS as parameters $\lambda_{gss}$ and $\lambda_{gst}$ vary. Higher bars indicate HOS closer to 1, signifying better domain adaptation performance. From Fig. \ref{fig4} we find that JAUA is sensitive to the parameters $\lambda_{gss}$ and $\lambda_{gst}$. The reason is that the parameter $\lambda_{gst}$ affects the importance of loss of the common class samples being misclassified as unknown class on the source domain while the parameter $\lambda_{gss}$ affects the importance of loss of the common class samples being misclassified as the private class on the source domain. Fig. \ref{fig4} shows that the HOS value increases as $\lambda_{gss}$ and $\lambda_{gst}$ increase. However, when $\lambda_{gst}=0.4$  and $\lambda_{gss}=0.1$, HOS drops dramatically. This occurs because this setup violates the restriction $\lambda_{gst}+\lambda_{gss} \leq m_{sk}/m_s$ in \textbf{Theorem 2}. On the other tasks of the other datasets, we find that JAUA is also sensitive to the other parameters. In real-world applications, we usually use the grid search to obtain the optimal hyperparameters.
\subsection{Convergence Analysis}
\begin{figure}[!t]
\centering
\includegraphics[width=3in]{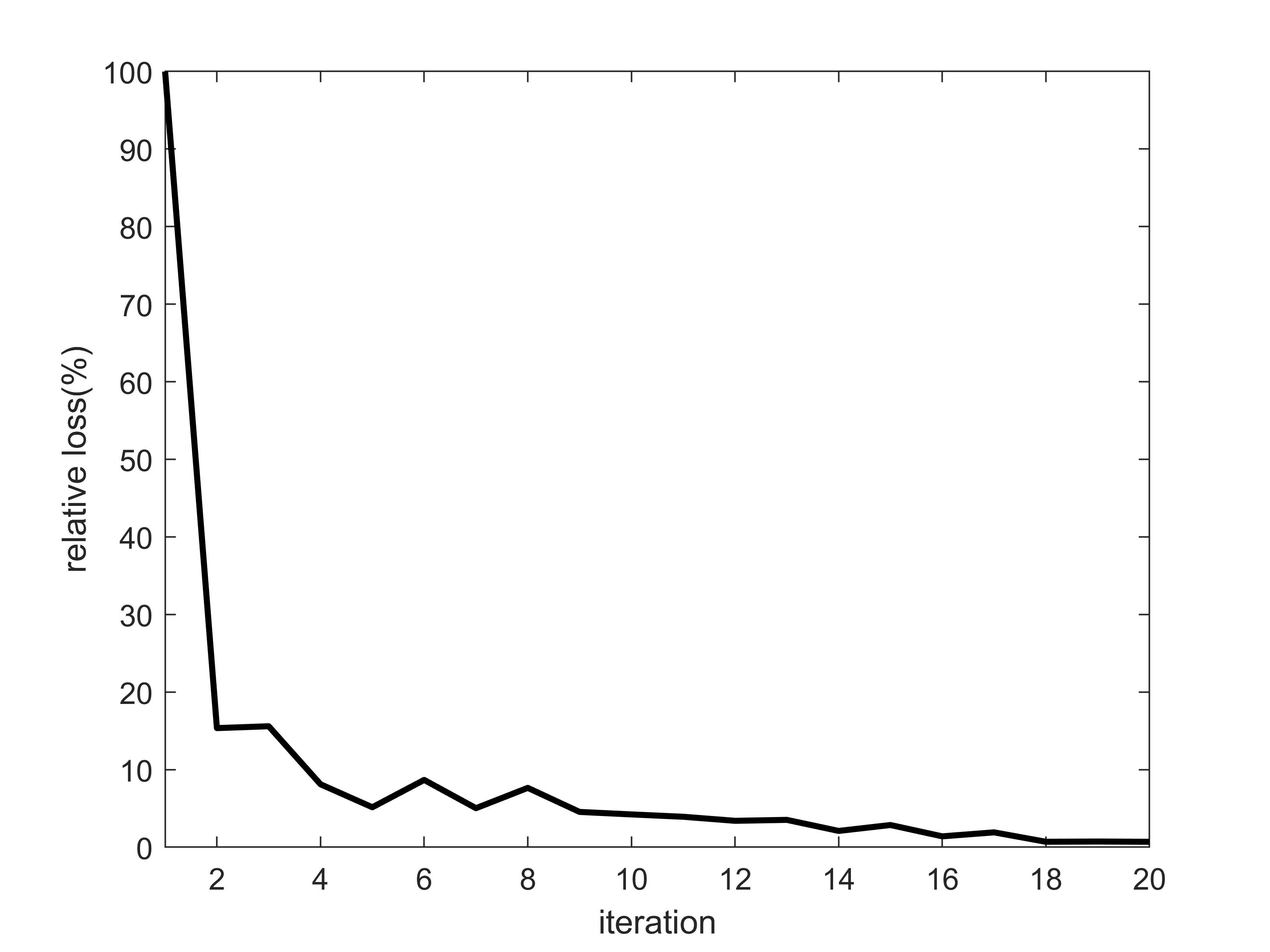}
\caption{The change of the whole loss of JAUA for task: D$\rightarrow$A on Office-31}
\label{fig5}
\end{figure}

In this subsection, we discuss the convergence of JAUA. Based on Algorithm \ref{algo1}, JAUA must set a maximum number of iterations $T$ to ensure the execution of the algorithm. In order to explore the convergence of the JAUA algorithm, a larger maximum number of iterations $T$ is determined in advance. The algorithm will stop when the relative loss change of two consecutive iterations is less than 1\%. Fig. \ref{fig5} shows the convergence of JAUA for task: D$\rightarrow$A of the dataset Office-31. We can also obtain similar conclusion for the other tasks on the other datasets.

\subsection{Ablation Study}
\begin{figure}[!t]
\centering
\includegraphics[width=3in]{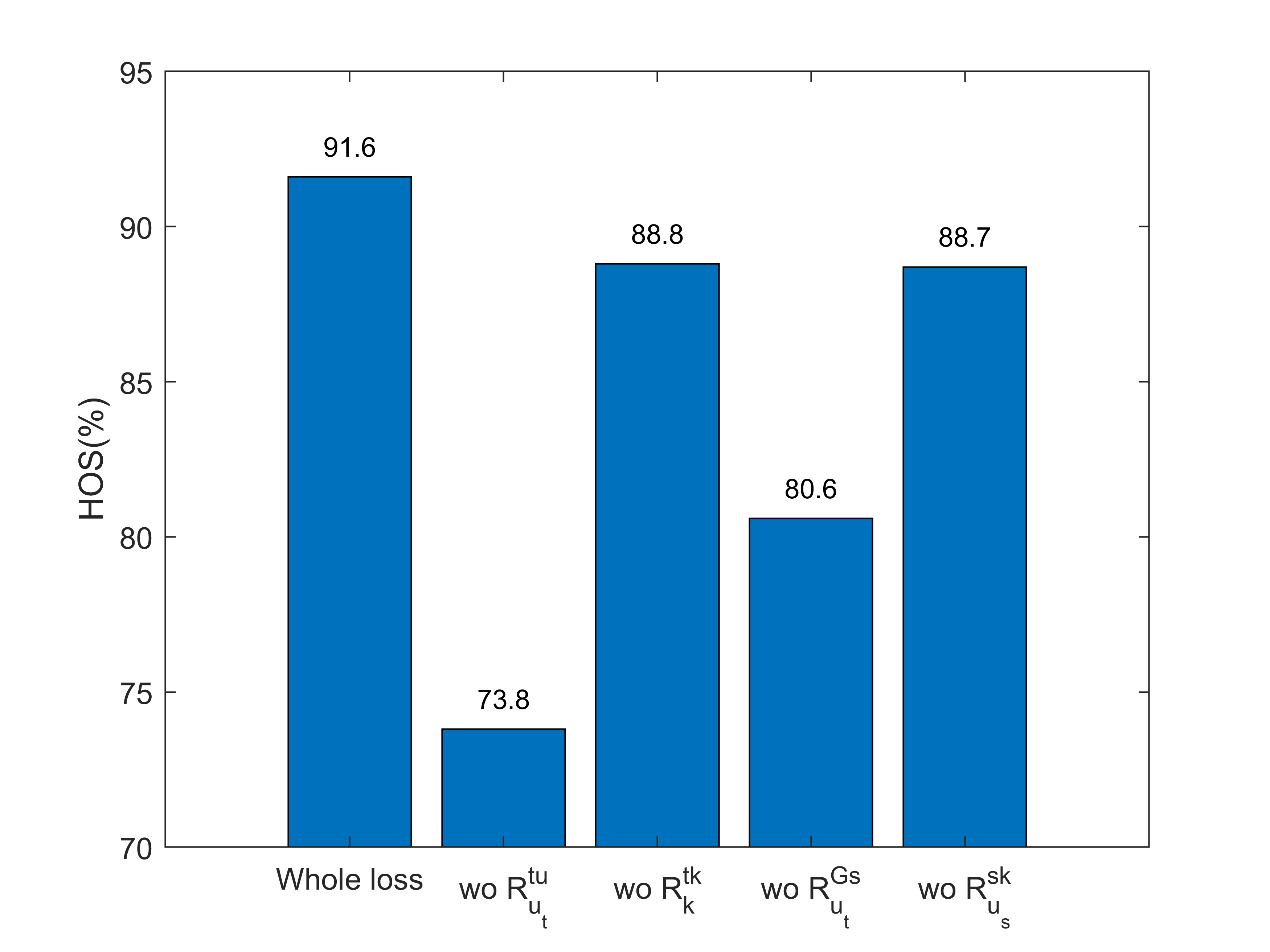}
\caption{Ablation experiment for task: D$\rightarrow$A on Office-31}
\label{fig6}
\end{figure}

To study the contribution of each component in the proposed model, we conduct the ablation experiment for task: D$\rightarrow$A on the Office-31 dataset. The results are illustrated in Fig. \ref{fig6}, where “wo X” means removing the loss term X from the whole loss function.

From Fig. \ref{fig6}, we can find that all of the components contribute to improving HOS of JAUA, where $R_{u_t}^{tu}$ and $R_{u_t}^{sk}$ are the two most important components for JAUA. This result implies that compared to UDA, the presence of unknown class samples makes it crucial to distinguish common and private class samples from unknown ones as much as possible, which in turn helps improve the performance of JAUA.

\section{Conclusions and Future Work}
In this paper, based on joint distribution alignment, a generalization error bound of the target domain is for the first time proposed for UniDA. A novel UniDA model is built by minimizing this bound. Based on a progressive pseudo-labeling strategy, JAUA is designed to identify known, private, and unknown class samples. The experiments show that JAUA can solve UniDA problem well.

In future work, we would like to further explore the intra-class structure of the unknown classes in the target domain. We expect to provide some novel error bounds and models by classifying the unknown class samples in the target domain into different unknown classes, which may be able to further improve the generalization ability of the model.

{\appendices
\section*{Appendix A. Proof of Theorem 1}
\textbf{\textit{Proof:}} 

{\color{black}We adopt an assumption below:

\textit{\textbf{Assumption 1}}\cite{wen2025open}: For any loss function $l$, $l$ is upper bounded, i.e. for any space $(\mathcal{X}*\mathcal{Y})$, $\exists M >0$, $\forall x \in \mathcal{X}$, $y\in \mathcal{Y}$, $l(h(x),y) \leq M$.

The specific definitions of $R^{tu}(h)$, $R^{tk}(h)$, $R^{su}(h)$ and $R^{sk}(h)$ are as follows:

\begin{equation}
R^{tu}(h)=\int_{\mathcal{X}*\mathcal{Y}^{u_t}}l(h(x),y)P_{X^tY^t}(x,y)dxdy\label{eq8}
\end{equation}

\begin{equation}
\begin{aligned}
R^{tk}(h)&=R^t(h)-R^{tu}(h)\\&=\int_{\mathcal{X}*\mathcal{Y}^{u_G}}l(h(x),y)P_{X^tY^t}(x,y)dxdy\label{eq9}
\end{aligned}
\end{equation}

\begin{equation}
R^{su}(h)=\int_{\mathcal{X}*\mathcal{Y}^{u_s}}l(h(x),y)P_{X^sY^s}(x,y)dxdy\label{eq10}
\end{equation}

\begin{equation}
\begin{aligned}
R^{sk}(h)&=R^s(h)-R^{su}(h)\\&=\int_{\mathcal{X}*\mathcal{Y}^{u_G}}l(h(x),y)P_{X^sY^s}(x,y)dxdy\label{eq11}
\end{aligned}
\end{equation}

Based on Assumption 1, according to \eqref{eq9} and \eqref{eq11}, we have 
\begin{equation}
\begin{aligned}
& \left|R^{t k}(h)-R^{s k}(h)\right| \\
& =\left|\int_{\mathcal{X}* \mathcal{Y}^{u_G}} l(h(x), y)\left[P_{X^t Y^t}(x, y)-P_{X^s Y^s}(x, y)\right] d x d y\right| \\
& \leq M\left|\int_{\mathcal{X}* \mathcal{Y}^{u_G}}\left[P_{X^t Y^t}(x, y)-P_{X^s Y^s}(x, y)\right] d x d y\right| \\
& \leq M D_{T V}\left(P_{X^s Y^s \sim \mathcal{X}* \mathcal{Y}^{u_G}} \| P_{X^t Y^t \sim \mathcal{X}* \mathcal{Y}^{u_G}}\right)\label{eq52}
\end{aligned}
\end{equation}
where $D_{TV}$ represents the Total Variation.

From \eqref{eq52}, we know that the following two inequalities hold:
\begin{equation}
\begin{aligned}
&R^{tk}(h) \\&\leq R^{sk}(h) + M D_{TV}\left(P_{X^s Y^s \sim \mathcal{X}* \mathcal{Y}^{u_G}} \| P_{X^t Y^t \sim \mathcal{X}* \mathcal{Y}^{u_G}}\right)\label{eq53}
\end{aligned}
\end{equation}

\begin{equation}
\begin{aligned}
&R^{sk}(h) \\&\leq R^{tk}(h) + M D_{TV}\left(P_{X^s Y^s \sim \mathcal{X}* \mathcal{Y}^{u_G}} \| P_{X^t Y^t \sim \mathcal{X}* \mathcal{Y}^{u_G}}\right)\label{eq54}
\end{aligned}
\end{equation}

Combining \eqref{eq4}, \eqref{eq5} with \eqref{eq8}, we can obtain
\begin{equation}
\begin{aligned}
&R^{t u}(h) \\
= & \int_{\mathcal{X} * \mathcal{Y}^{u_t}} l(h(x), y) P_{X^t Y^t}(x, y) d x d y \\
= & R_{u_t}^t(h)-R_{u_t}^{sk}(h)+ \\
& \int_{\mathcal{X} * \mathcal{Y}^{u_G}} l\left(h(x), y^{u_t}\right)\left[P_{X^s Y^s}(x, y)-P_{X^t Y^t}(x, y)\right] d x d y \\
\leq & \lambda_{t t} R_{u_t}^t(h)-\lambda_{g s t} R_{u_t}^{sk}(h) \\
& +N_T D_{T V}\left(P_{X^s Y^s \sim \mathcal{X}* \mathcal{Y}^{u_G}} \| P_{X^t Y^t \sim \mathcal{X}* \mathcal{Y}^{u_G}}\right)\label{eq55}
\end{aligned}
\end{equation}
where $\lambda_{gst} \leq 1$ and $\lambda_{tt} \geq 1$ are two hyperparameters which control the importance of $R_{u_t}^{sk} (h)$ and $R_{u_t}^t (h)$.

Combining \eqref{eq9}, \eqref{eq53} with \eqref{eq55}, we can obtain
\begin{equation}
\begin{aligned}
R^t(h) =& R^{tu}(h) + R^{tk}(h) \\
\leq & \lambda_{tt} R^t_u(h) - \lambda_{gst} R^{sk}_{u_t}(h) \\
& + N_T D_{TV}\left(P_{X^s Y^s \sim \mathcal{X}* \mathcal{Y}^{u_G}} \big\| P_{X^t Y^t \sim \mathcal{X}* \mathcal{Y}^{u_G}} \right)+ \\
& R^{sk}(h)+M D_{TV}\left(P_{X^s Y^s \sim \mathcal{X}* \mathcal{Y}^{u_G}} \big\| P_{X^t Y^t \sim \mathcal{X}* \mathcal{Y}^{u_G}} \right) \\
= &\lambda_{tt} R^t_{u_t}(h) - \lambda_{gst} R^{sk}_{u_t}(h) + R^{sk}(h) \\
& + (M + N_T) D_{TV}\left(P_{X^s Y^s \sim \mathcal{X}* \mathcal{Y}^{u_G}} \big\| P_{X^t Y^t \sim \mathcal{X}* \mathcal{Y}^{u_G}} \right)\label{eq56}
\end{aligned}
\end{equation}
Similarly, combining \eqref{eq6}, \eqref{eq12} with \eqref{eq10}, we have
\begin{equation}
\begin{aligned}
&R^{s u}(h) \\
= & \int_{\mathcal{X} * \mathcal{Y}^{u_s}} l(h(x), y) P_{X^s Y^s}(x, y) d x d y \\
= & R_{u_s}^s(h)-R_{u_s}^{tk}(h) \\
& -\int_{\mathcal{X} * \mathcal{Y}^{u_G}} l\left(h(x), y^{u_s}\right)\left[P_{X^s Y^s}(x, y)-P_{X^t Y^t}(x, y)\right] d x d y \\
\geq & \lambda_{g s s} R_{u_s}^s(h)-R_{u_s}^{tk}(h) \\
& -N_S D_{T V}\left(P_{X^s Y^s \sim \mathcal{X}* \mathcal{Y}^{u_G}} \| P_{X^t Y^t \sim \mathcal{X}* \mathcal{Y}^{u_G}}\right)\label{eq57}
\end{aligned}
\end{equation}
where $\lambda_{gss} \leq 1$ is a hyperparameter which controls the importance of $R^s_{u_s}(h)$.

Combining \eqref{eq11}, \eqref{eq53} with \eqref{eq57}, we have
\begin{equation}
\begin{aligned}
&R^{tk}(h) \\
\leq &R^{sk}(h)+MD_{TV}(P_{X^s Y^s \sim \mathcal{X}* \mathcal{Y}^{u_G}} \| P_{X^t Y^t \sim \mathcal{X}* \mathcal{Y}^{u_G}}) \\
\leq &R^{sk}(h)+MD_{TV}(P_{X^s Y^s \sim \mathcal{X}* \mathcal{Y}^{u_G}} \| P_{X^t Y^t \sim \mathcal{X}* \mathcal{Y}^{u_G}}) \\
& +R^{su}(h)-\lambda_{gss}R^s_{u_s}(h)+R^{tk}_{u_s}(h) \\
& +N_SD_{TV}(P_{X^s Y^s \sim \mathcal{X}* \mathcal{Y}^{u_G}} \| P_{X^t Y^t \sim \mathcal{X}* \mathcal{Y}^{u_G}}) \\
= &R^s(h)+R^{tk}_{u_s}(h)-\lambda_{gss}R^s_{u_s}(h) \\
& +(M+N_S)D_{TV}(P_{X^s Y^s \sim \mathcal{X}* \mathcal{Y}^{u_G}} \| P_{X^t Y^t \sim \mathcal{X}* \mathcal{Y}^{u_G}})\label{eq58}
\end{aligned}
\end{equation}

From \cite{sason2016f}, we know that the following inequality holds:
\begin{equation}
\begin{aligned}
&D_{TV}(P_{X^s Y^s \sim \mathcal{X}* \mathcal{Y}^{u_G}} \| P_{X^t Y^t \sim \mathcal{X}* \mathcal{Y}^{u_G}}) \\&
\leq \sqrt{{2 D_{CS}(P_{X^s Y^s \sim \mathcal{X}* \mathcal{Y}^{u_G}} \| P_{X^t Y^t \sim \mathcal{X}* \mathcal{Y}^{u_G}}})}\label{eq59}
\end{aligned}
\end{equation}

Using \eqref{eq54},\eqref{eq56},\eqref{eq58} and \eqref{eq59}, we can obtain
\begin{equation}
\begin{aligned}
& R^t(h) \\
\leq & \lambda_{tt}R^t_{u_t}(h)-\lambda_{gst}R^{sk}_{u_t}(h)+R^{sk}(h) \\
& +(M+N_T)D_{TV}(P_{X^s Y^s \sim \mathcal{X}* \mathcal{Y}^{u_G}} \| P_{X^t Y^t \sim \mathcal{X}* \mathcal{Y}^{u_G}}) \\
\leq & \lambda_{tt}R^t_{u_t}(h)-\lambda_{gst}R^{sk}_{u_t}(h)+R^{tk}(h) \\
& +(2M+N_T)D_{TV}(P_{X^s Y^s \sim \mathcal{X}* \mathcal{Y}^{u_G}} \| P_{X^t Y^t \sim \mathcal{X}* \mathcal{Y}^{u_G}}) \\
\leq & R^s(h)+\lambda_{tt}R^t_{u_t}(h)-\lambda_{gst}R^{sk}_{u_t}(h)\\
&+R^{tk}_{u_s}(h)-\lambda_{gss}R^s_{u_s}(h)+(3M+N_T+N_S) \\
& *\sqrt{{2 D_{CS}(P_{X^s Y^s \sim \mathcal{X}* \mathcal{Y}^{u_G}} \| P_{X^t Y^t \sim \mathcal{X}* \mathcal{Y}^{u_G}}})}\label{eq60}
\end{aligned}
\end{equation}

As stated in the Problem Setting, the goal of UniDA is to minimize $R^t(h)$. Based on \eqref{eq4} and \eqref{eq12}, we know that in order to minimize $R^t (h)$, we need to minimize $R ^s (h)$, $R_{u_t}^{tu} (h)$, $R_{u_t}^{tk} (h)$ and $R_{u_s}^{tk} (h)$ and maximize $R_{u_t}^{sk} (h)$, $R_{u_s}^{su} (h)$ and $R_{u_s}^{sk} (h)$. Generally speaking, minimizing $R^s (h)$ and $R_{u_t}^{tu} (h)$ can enable the classifier to more accurately classify source samples and unknown class samples. Similarly, maximizing $R_{u_t}^{sk} (h)$ and $R_{u_s}^{sk}(h)$ can enable the classifier to more accurately classify the common class samples. Unfortunately, minimizing $R^{tk}_{u_t}(h)$ and $R^{tk}_{u_s}(h)$ may result in that the classifier classifies the common class samples from the target domain into unknown class or private classes. In the same way, maximizing $R^{su}_{u_s}(h)$ may result in that the classifier  classifies the private class samples from the source domain into common classes.} In order to avoid these cases, in the following, we update the upper bound in \eqref{eq60}. Based on the law of large numbers, we can easily find that $R^{tk}_{u_t}(h)$ and $R^{tk}_{u_s}(h)$ are finite, there exist positive real numbers $A$ and $B$ which satisfy $R^{tk}_{u_t}(h) \leq \frac{\lambda_t}{\lambda_{tt}}R^{tk}_k(h)+\frac{A}{\lambda_{tt}}$ and $R^{tk}_{u_s}(h) \leq \lambda_s R^{tk}_k(h)+B$, where $\lambda_t$ and $\lambda_s$ are hyperparameters which bound $R^{tk}_{u_t}(h)$ and $R^{tk}_{u_s}(h)$.

Based on the above analysis, the novel bound of $R^t (h)$ can be rewritten as: 
\begin{equation}
\begin{aligned}
& R^t(h) \\
\leq & R^s(h)+\lambda_{tt}R^t_{u_t}(h)-\lambda_{gst}R^{sk}_{u_t}(h)\\
&+R^{tk}_{u_s}(h)-\lambda_{gss}R^s_{u_s}(h)+(3M+N_T+N_S) \\
& *\sqrt{{2 D_{CS}(P_{X^s Y^s \sim \mathcal{X}* \mathcal{Y}^{u_G}} \| P_{X^t Y^t \sim \mathcal{X}* \mathcal{Y}^{u_G}}})} \\
= &R^s(h)+\lambda_{tt}R^{tu}_{u_t}(h)+\lambda_{tt}R^{tk}_{u_t}(h)-\lambda_{gst}R^{sk}_{u_t}(h)+R^{tk}_{u_s}(h)\\
&-\lambda_{gss}R^{su}_{u_s}(h)-\lambda_{gss}R^{sk}_{u_s}(h)+ (3M+N_T+N_S)\\
& *\sqrt{{2 D_{CS}(P_{X^s Y^s \sim \mathcal{X}* \mathcal{Y}^{u_G}} \| P_{X^t Y^t \sim \mathcal{X}* \mathcal{Y}^{u_G}}})}\\
\leq &R^s(h)+\lambda_{tt}R^{tu}_{u_t}(h)+(\lambda_t+\lambda_s)R^{tk}_k(h)\\
& -\lambda_{gst}R^{sk}_{u_t}(h)-\lambda_{gss}R^{sk}_{u_s}(h)+(3M+N_T+N_S) \\
& *\sqrt{{2 D_{CS}(P_{X^s Y^s \sim \mathcal{X}* \mathcal{Y}^{u_G}} \| P_{X^t Y^t \sim \mathcal{X}* \mathcal{Y}^{u_G}}})}+A+B\label{eq15}
\end{aligned}
\end{equation}

Therefore, \textbf{Theorem 1} has been proved.}

\section*{Appendix B. Proof of Theorem 2}

\textbf{\textit{Proof:}} {\color{black}Let
\begin{equation}
\begin{aligned}
&\mathcal{L}(\alpha)\\ 
&=\min_{\alpha \in \mathcal{R}^{(m_s+m_t)*(C+2)}} \eta tr(\alpha^\top K \alpha)+\|(Y-\alpha^\top K)W\|^2_2 \\&+\lambda_{tt}\|(Y_{u_t}-\alpha^\top K)V_{tu}\|^2_2+ (\lambda_{s}+\lambda_{t})\|(Y_{k}-\alpha^\top K)V_{k}\|^2_2 \\
&- \lambda_{gst}\|(Y_{u_t}-\alpha^\top K)V_{sk}\|^2_2 -\lambda_{gss}\|(Y_{u_s}-\alpha^\top K)V_{sk}\|^2_2 
\label{eq61}
\end{aligned}
\end{equation}
if and only if $\frac{\partial^2 \mathcal{L}(\alpha)}{\partial \alpha^2} \geq 0$,  the solution of $\frac{\partial \mathcal{L}(\alpha)}{\partial \alpha}=0$ can make $\mathcal{L}(\alpha)$ to reach minimum.

From
\begin{equation}
\begin{aligned}
\frac{\partial \mathcal{L}(\alpha)}{\partial \alpha}=&-2KW^2Y^\top+2KW^2K\alpha-2\lambda_{tt}KV_{tu}^2Y_{u_t}^\top\\
&+2\lambda_{tt} KV_{tu}^2K\alpha-2(\lambda_s+\lambda_t)KV_k^2 Y_k^\top\\
&+2(\lambda_s+\lambda_t)KV_k^2K\alpha+2\lambda_{gst}KV_{sk}^2Y_{u_t}^\top\\
&-2\lambda_{gst}KV_{sk}^2 K\alpha+2\lambda_{gss}KV_{sk}^2Y_{u_s}^\top\\
&-2\lambda_{gss}KV_{sk}^2K\alpha+2\eta K \alpha
\label{eq62}
\end{aligned}
\end{equation}
we have
\begin{equation}
\begin{aligned}
\frac{\partial^2 \mathcal{L}(\alpha)}{\partial \alpha^2}=&2(I\otimes KW^2K)+2\lambda_{tt}(I\otimes KV_{tu}^2K)\\
&+2(\lambda_s+\lambda_t)(I\otimes KV_k^2K)-2\lambda_{gst}(I\otimes KV^2_{sk}K)\\
&-2\lambda_{gss}(I\otimes KV^2_{sk}K)+2\eta(I\otimes K) \label{eq63}
\end{aligned}
\end{equation}
where $\otimes$ represents Kronecker product. Based on \eqref{eq23} and \eqref{eq26}, we can find that $\frac{\partial^2 \mathcal{L}(\alpha)}{\partial \alpha^2} \geq 0$ always holds if and only if $\frac{1}{m_s}-\frac{\lambda_{gst}+\lambda_{gss}}{m_{sk}} \geq 0$. Hence, \textbf{Theorem 2} has been proved.}

\bibliographystyle{IEEEtran}
\bibliography{IEEEabrv,references}

\newpage

 




\vfill
\end{document}